%% file: main.tex
\documentclass[10pt,twocolumn,letterpaper]{article}

\usepackage[pagenumbers]{cvpr} 

\usepackage{tcolorbox}

\definecolor{cvprblue}{rgb}{0.21,0.49,0.74}

\usepackage[
    pagebackref,
    breaklinks,
    colorlinks,
    linkcolor=red,     
    citecolor=cvprblue,         
    urlcolor=magenta        
]{hyperref}

\def\paperID{****} 
\def\confName{****}
\def\confYear{****}

\title{ChainReaction: Causal Chain-Guided Reasoning for Modular and Explainable Causal-Why Video Question Answering}

\author{Paritosh Parmar$^{1,2}$\thanks{Equal contribution.} \hspace{1cm} Eric Peh$^{1,2}$\footnotemark[1] \hspace{1cm} Basura Fernando$^{1,2,3}$\\
\small{$^{1}$Institute of High Performance Computing, Agency for Science, Technology and Research, Singapore}\\
\small{$^{2}$Centre for Frontier AI Research, Agency for Science, Technology and Research, Singapore}\\
\small{$^{3}$College of Computing and Data Science, Nanyang Technological University, Singapore}\\
}

\begin{document}
\maketitle

\input{Sections/Abstract}
\input{Sections/Introduction}
\input{Sections/Related}
\input{Sections/Dataset}
\input{Sections/Approach}
\input{Sections/Experiments}
\input{Sections/Conclusion}

\paragraph{Acknowledgements.} This research/project is supported by the National Research Foundation, Singapore, under its NRF Fellowship (Award\# NRF-NRFF14-2022-0001). This research is also supported by funding allocation to B.F. by the Agency for Science, Technology and Research (A*STAR) under its SERC Central Research Fund (CRF), as well as its Centre for Frontier AI Research (CFAR). We would also like to thank Dhruv Verma and Elston Tan.

{
    \small
    \bibliographystyle{ieeenat_fullname}
    \bibliography{main}
}



\end{document}

%% file: Sections/Abstract.tex
\begin{abstract}
    Existing Causal-Why Video Question Answering (VideoQA) models often struggle with higher-order reasoning, relying on opaque, monolithic pipelines that entangle video understanding, causal inference, and answer generation. These black-box approaches offer limited interpretability and tend to depend on shallow heuristics. We propose a novel, modular paradigm that explicitly decouples causal reasoning from answer generation, introducing natural language causal chains as interpretable intermediate representations. Inspired by human cognitive models, these structured cause-effect sequences bridge low-level video content with high-level causal reasoning, enabling transparent and logically coherent inference. Our two-stage architecture comprises a Causal Chain Extractor (CCE) that generates causal chains from video-question pairs, and a Causal Chain-Driven Answerer (CCDA) that derives answers grounded in these chains. To address the lack of annotated reasoning traces, we introduce a scalable method for generating accurate causal chains from existing datasets. We construct human verified causal chains for 46K samples. We also propose CauCo, a new evaluation metric for causality-oriented captioning. Experiments on three large-scale benchmarks demonstrate that our approach not only outperforms state-of-the-art models, but also yields substantial gains in explainability, user trust, and generalization—positioning the CCE as a reusable causal reasoning engine across diverse domains.
\end{abstract}

%% file: Sections/Introduction.tex
\section{Introduction}
\label{sec:introduction}

Understanding the motivations behind human actions is crucial for developing advanced systems for nuanced behavior analysis. Human actions are shaped by factors such as personal experience, emotion, social context, and culture. This complexity requires uncovering underlying causes. In this context, Causal Video Question Answering (Causal-Why VideoQA) asks models not only to recognize events but to explain why they occur—demanding higher-order reasoning beyond descriptive QA~\cite{causalvidqa,nextqa,causalchaos,foss2025causalvqa}.

Existing Causal-Why VideoQA models often reason from incomplete evidence or rely on shallow heuristics (\eg, matching action verbs or object nouns in vision-language embedding spaces \cite{causalchaos, rawal2024dissecting, wei2023visual}). These models entangle video understanding, reasoning, and answer generation into one monolithic process, making their reasoning opaque and error-prone. Many high-performing vision language models (VLMs) also operate as black boxes, offering limited interpretability into their decisions.

\begin{figure}
    \centering
    \small
    \includegraphics[width=\columnwidth]{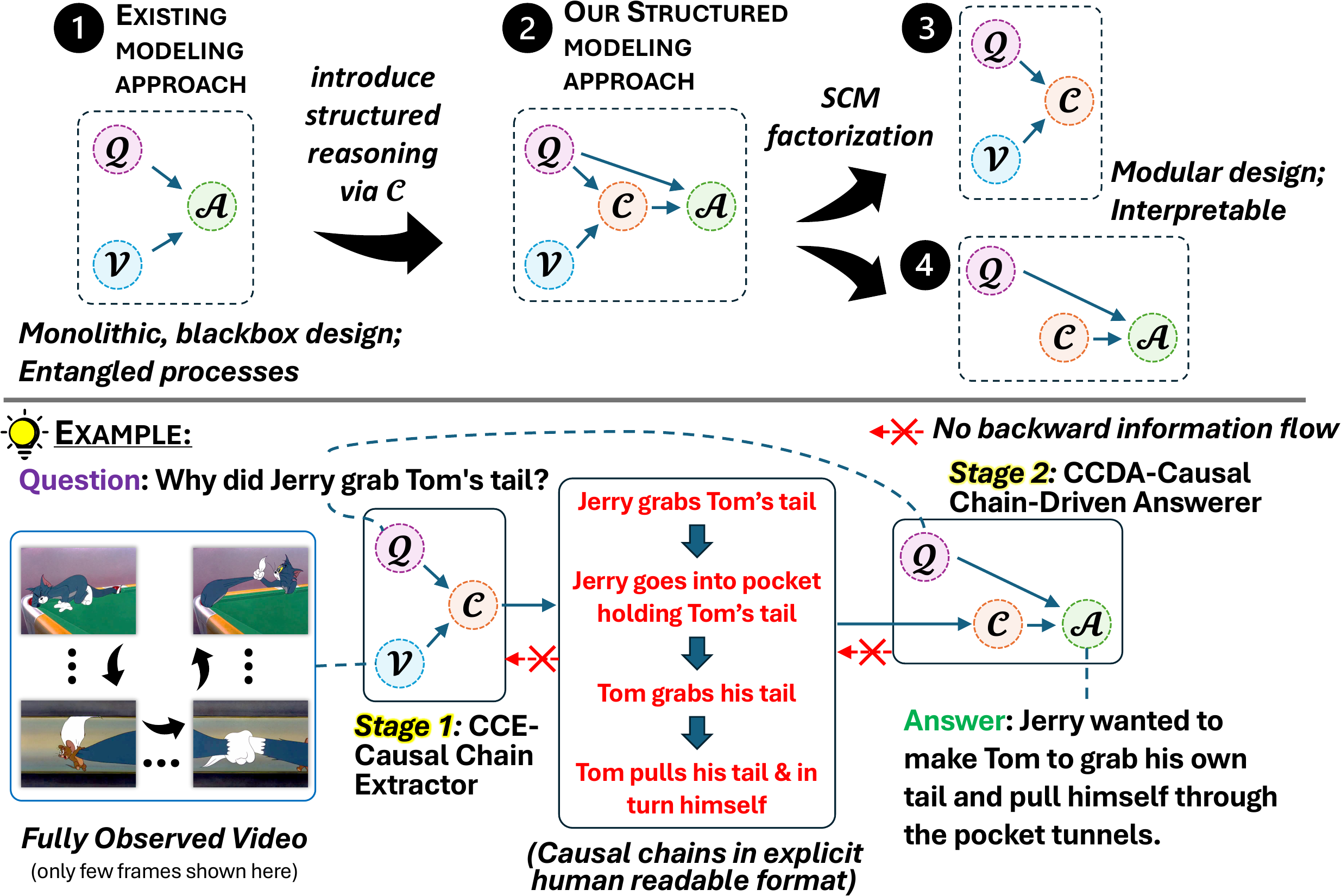}
    \caption{\textbf{(Top)~Concept.} 
     \textbf{(1)} Existing Video ($\mathcal{V}$) Question ($\mathcal{Q}$) Answer ($\mathcal{A}$) approaches through the lens of structural causal models (SCMs), highlighting their monolithic and black-box nature. \textbf{(2)} In contrast, we propose a principled departure from this paradigm: leveraging the Causal Reasoning Trace ($\mathcal{C}$), a structured intermediate representation based on natural language causal chains. We factorize this SCM into two SCMs \textbf{(3,4)}---enabling structured video understanding, reasoning, and inference---leading to superior explainability and performance. \textbf{(Bottom) Example.} \textit{Please zoom in for the best view.}
     }
    \label{fig:concept}
\end{figure}

In this work, we make the case that reasoning and answering should be explicitly decoupled and modularized. We introduce a new paradigm in which causal reasoning and answer generation are handled by separate modules that communicate via natural language causal chains~\cite{hanson1955causal, pearl2018book}---structured sequences of cause-effect events that serve as intermediate reasoning steps (\autoref{fig:concept}). 
Causal chains or causal reasoning traces provide a logically coherent bridge between low-level video content and high-level causal understanding~\cite{trabasso1985causal}. Formulated as natural-language sequences, they capture the linear, observable steps linking a cause (\eg, a character’s intention) to its effect, ensuring that answers are grounded in the video's causal progression.
To our knowledge, ours is the first approach in Causal-Why VideoQA to explicitly use causal chains as interpretable intermediates, inspired by their role in human cognition and scientific explanation~\cite{gopnik2004theory}. 

At the core of our method lies the integration of Structural Causal Models (SCMs) and Chain-of-Thought (CoT) reasoning, which enables a principled, structured decomposition of the VideoQA task. This synergy allows us to model causality in a robust and interpretable way, going beyond latent embeddings to explicitly capture causal semantics.

Our model consists of two stages: \textbf{1)} a causal chain extractor---CCE---(\autoref{fig:concept}(3)), and \textbf{2)} a causal chain-driven answerer---CCDA---(\autoref{fig:concept}(4)). The CCE model learns to extract causal chains from video, conditioned on the causal-why questions. The CCDA model learns to generate answers to causal-why questions based on the extracted causal chains. 
Distinct from general causal inference models that reason over partially observed or hypothetical systems, the CCE operates in fully observed video settings, where the causal process has already unfolded. After observing the complete video, the CCE identifies the actualized, linear sequence of events that led to the outcome—making the task descriptive and post-hoc rather than inferential, and emphasizing interpretability and causal fidelity within the VideoQA framework. 
However, a key challenge lies in training CCE. In particular, there are no datasets containing  reasoning traces for training CCE. To tackle this challenge, we develop an approach to generate causal chains from existing VideoQA datasets efficiently.

Extensive experiments on three large scale datasets demonstrate: \textbf{1)} causal chains are promising intermediate representations; \textbf{2)} performance improvements across all three datasets; \textbf{3)} human studies showed that causal chain-driven video QA enhances explainability and interpretability from multiple perspectives; \textbf{4)} the causal chain extractor generalizes well to out-of-domain datasets, highlighting its potential as a reusable causal reasoning engine.

Our main contributions can be summarized as: 
\begin{itemize}
\item Proposing a structured paradigm for Causal-Why Video Question Answering that leverages natural language causal chains as intermediate representations to enhance reasoning and transparency.
\item Introducing a two-stage architecture—Causal Chain Extractor (CCE) and Causal Chain-Driven Answerer (CCDA)—that decouples video understanding from causal inference.
\item Introducing a human-in-the-loop framework that uses large language models to propose causal chain drafts, which are subsequently verified and finalized by human annotators, yielding high-quality causal reasoning data.
\item Introducing CauCo score, a causality-oriented captioning metric.
\item Demonstrating through extensive experiments and human studies that the proposed approach outperforms state-of-the-art models while offering significant gains in explainability, user trust, and system debuggability.
\item Showing that the CCE generalizes well to out-of-domain datasets, enabling effective causal reasoning across diverse video domains.
\end{itemize}

%% file: Sections/Related.tex
\section{Related Work}
\label{sec:related}

\noindent\textbf{Chain-of-Thought (CoT)} is a prompting technique that improves LLMs by decomposing tasks into intermediate reasoning steps. It is effective in arithmetic, logic, and commonsense tasks, especially via few-shot prompting with exemplars \cite{wei2022chain}. Unlike CoT methods that treat reasoning as emergent, we model reasoning steps—causal chains—as explicit, structured variables. This enables supervision, interpretability, and integration with modular architectures, breaking from the monolithic CoT paradigm.

\noindent\textbf{Causal Reasoning and Structural Causal Models (SCMs).}
Causality is foundational in scientific reasoning and is formalized in AI through SCMs and do-calculus~\cite{pearl2009causality}. Recent work emphasizes causal representation learning for robustness and generalization~\cite{scholkopf2021toward}, and SCM-based methods have been used for tasks like debiasing~\cite{tang2020unbiased} and counterfactual explanations~\cite{narendra2018explaining}. However, such approaches are rarely applied to VideoQA. We address this gap by integrating SCMs into a modular framework where video-derived causal chains support interpretable answer generation—bringing structured causal reasoning to a domain where temporal complexity often obscures transparency.

\noindent\textbf{Causal Chains and Structured Reasoning in AI.}
Causal chains provide intuitive, stepwise explanations of how events unfold~\cite{causalchaos,niu2025r}. In AI, structured representations like causal graphs improve reasoning and interpretability, and prior work has explored causal interactions in videos or intervention-based action recognition~\cite{wang2024modeling,ayazoglu2013finding}. However, these approaches do not use causal chains as intermediates for reasoning. We instead formalize causal chains as explicit bridges between video observations and high-level causal understanding, enabling a principled and interpretable framework for Causal-Why Video QA.

\noindent\textbf{Video Causal Reasoning \& Temporal Understanding.} 
Recent work explores causality in video understanding—\eg, structural models for causal interactions and moment retrieval~\cite{li2020causal,yang2021deconfounded}—but not for QA. Other methods select key segments for QA~\cite{wei2023visual} or model object interactions generatively~\cite{chen2023llcp}, yet their causal reasoning remains implicit or localized.
Chen \etal~\cite{chen2025mecd} introduce event-level causal diagram annotation, but their models explain a final event rather than generate causal chains or use them for QA. In contrast, we extract event-level causal chains and leverage them in a dedicated inference module.
Zang \etal~\cite{zang2023discovering} study video–text causal links, and Su \etal~\cite{su2023language} generate causal questions from captions, but neither models causal chains or decouples reasoning from answering.

\noindent\textbf{Vision-Language Models (VLMs) \& their Limitations in Causality.}
Recent VLMs (\eg, VideoLLaMA~\cite{videollama}, VideoChat2~\cite{videochat2},  VILA~\cite{vila}, \etc) advance VideoQA \& captioning through largescale video–text pretraining, but primarily recognize and describe events rather than explain them. Their monolithic architectures offer limited interpretability \& often rely on shallow correlations. Our method complements them by introducing causal chains as structured reasoning intermediates, improving both performance \& transparency in Causal-Why Video QA.

%% file: Sections/Dataset.tex
\section{Causal Chain Construction for SFT}
\label{sec:dataset}

We propose a novel, scalable and efficient methodology to construct causal chains to be used for supervised finetuning (SFT) of our causal chain extractor (\autoref{sec:approach}). We generate causal chains that accurately reflect the reasoning behind human-written answers to video-based questions. \\
\textbf{Base datasets:} since no existing VideoQA dataset includes causal chain annotations to support our approach, we collect causal chains for three challenging causal video QA datasets: \textbf{1)} \textit{NextQA} \cite{nextqa}, \textbf{2)} \textit{CausalVidQA} \cite{causalvidqa}, and \textbf{3)} \textit{CausalChaos!} \cite{causalchaos}. We refer to these three datasets as the base (or source) datasets. We focus on \textit{Causal-Why QA} in these base datasets. In base datasets, each sample is a paired triplet of \{Video ($\mathcal{V}$), Question ($\mathcal{Q}$), Answer ($\mathcal{A}$)\}. \\
\textbf{Causality defined in the base datasets:} the causal-why questions and correct answers in base datasets are human-authored, as a result the notion of cause or causality is well defined in the data itself. All the annotations are rigorously cross-annotator verified, resulting in multihuman agreement on causality. Our goal is to learn to reason through these causal relations. We have provided further details on these three base datasets in the \textcolor{RoyalBlue}{Appendix}. 

\vspace{0.1cm}

\begin{figure}
    \centering
    \includegraphics[width=\columnwidth]{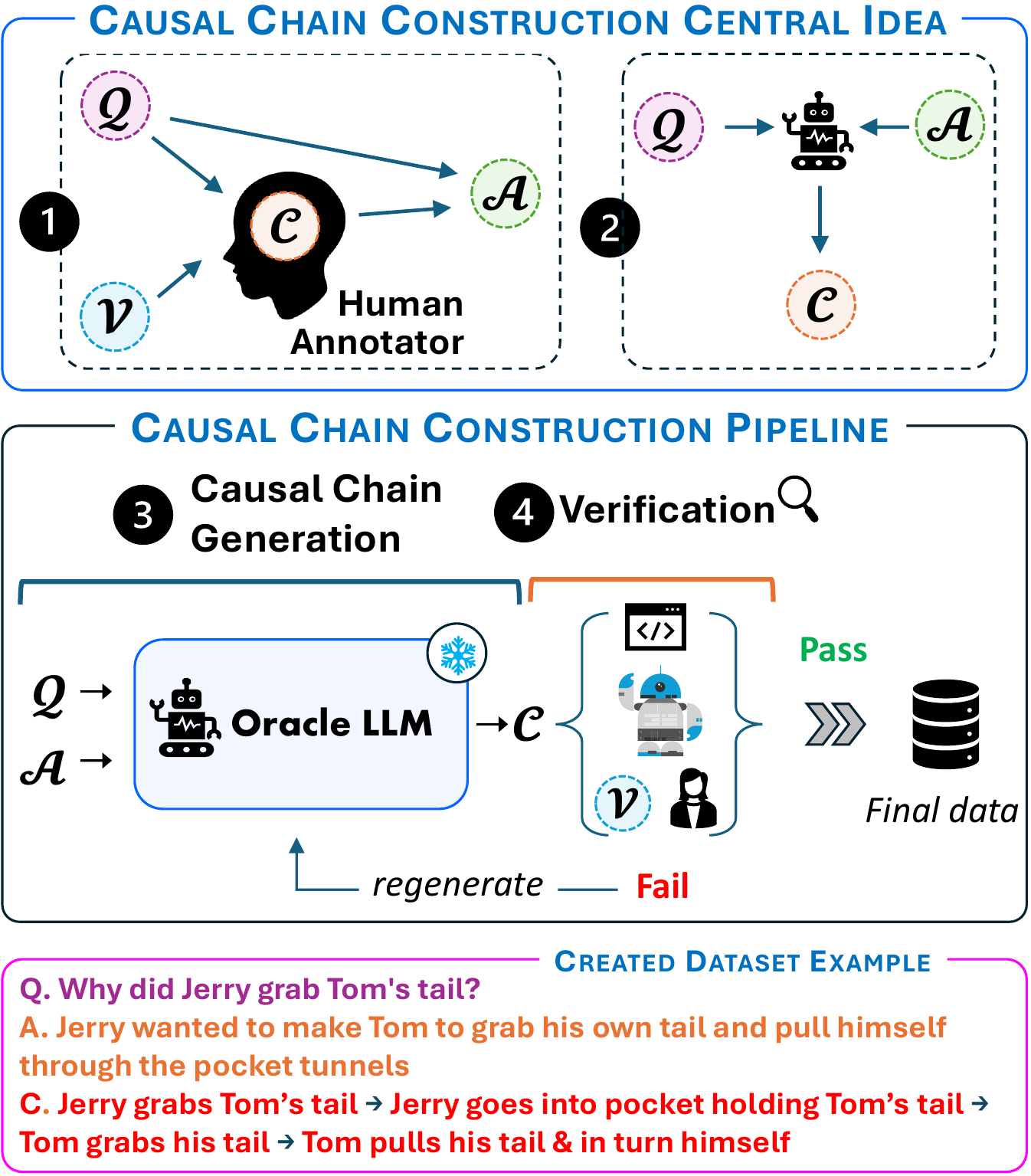}
    \caption{\textbf{Causal chain construction for SFT.} \textbf{(1)} Human annotators of base datasets intuitively and implicitly make use of causal chains when writing correct answers. \textbf{(2)} We propose to recover these causal chains with the help of LLM using questions and correct gold answers. \textbf{(3,4)} Our robust causal chain generation and manual verification and video grounding check pipeline.}
    \label{fig:dataset_generation}
\end{figure}

\noindent\textbf{Preliminary I.} 
Human annotators watch the video ($\mathcal{V}$), formulate a question ($\mathcal{Q}$) about it, and write the correct/gold answer ($\mathcal{A}$). The annotations are cross-verified, ensuring multi-annotator agreement. Annotators implicitly use causal chains ($\mathcal{C}$) when writing answers, as illustrated in \autoref{fig:dataset_generation}(1). Thus, the detailed QA pairs of our base datasets contain causal chains embedded within them.

\vspace{0.1cm}

\noindent\textbf{Preliminary II.} Given the questions and the corresponding correct, gold answers, LLMs have been demonstrated to accurately recover the intermediate reasoning steps or causal chains. Illustrated in \autoref{fig:dataset_generation}(2).

\vspace{0.1cm}

\noindent\textbf{Corollary I.} Based on preliminaries I and II, we propose that if provided with the questions and the corresponding human written correct, gold answers from the base datasets, then powerful oracle LLMs (\eg, \cite{gpt4o, deepseekvl2, gemini}) can reliably recover the reasoning steps or causal chains. 

\vspace{0.1cm}

\noindent\textbf{Causal Chain Construction Methodology.} Following Corollary I, we propose a robust methodology for causal chain generation and verification in the following.
\begin{enumerate}[wide, labelwidth=!, labelindent=0pt]
    \item \textbf{Causal Chain Draft Generation Process:} We leverage a powerful LLM (GPT4o). We refer to it as Oracle LLM. We prompt the Oracle LLM with question ($\mathcal{Q}$) and the corresponding human written correct, gold answer ($\mathcal{A}$) and ask it to generate a causal chain ($\mathcal{C}$) in natural language in a specified format. Specifically, we instruct the LLM to return causal chains in a structured format: \texttt{[Event A] → [Event B] → [Event C]...}\footnote{Note that we do not train the LLMs to generate causal chains; rather, directly capitalize on their inherent knowledge, commonsense reasoning \& synthesis capabilities to derive them.} Full prompt provided in the \textcolor{RoyalBlue}{Appendix}. Illustrated in \autoref{fig:dataset_generation}(3). Oracle LLM is not used as end-task solvers but as structured reasoning components within a rigorously controlled pipeline. Their role is analogous to that of annotator assistants---providing drafts that undergo multi-layer human validation to ensure reliability and reproducibility.

    \item \textbf{Causal Chain Verification Process:} Illustrated in \autoref{fig:dataset_generation}(4). We ensure that the generated causal chains are of high quality and accurately capture the reasoning steps using rigorous Quality Checks consisting of:  
    \begin{enumerate}[wide, labelwidth=!, labelindent=0pt, noitemsep,topsep=0pt]
        \item \textbf{\textit{Programmatic Validation:}} 
        Programs check generated causal chains for structure, format, completeness, and length ($\leq$10 events).
        
        \item \textbf{\textit{Cross-LLM Verification:}} Verifier LLM is provided with question ($\mathcal{Q}$), correct gold answer ($\mathcal{A}$) and the generated causal chain ($\mathcal{C}$) Generated causal chains are independently reviewed and verified by second LLM to assess the chain's logical coherence, relevance, and consistency with $\mathcal{Q}$ and $\mathcal{A}$ pair. For this, we use a powerful LLM, but different than Oracle LLM (GPT4o) to avoid LLM circularity bias. Specifically, we use Gemini 2.5 as the verifier LLM. 
        
        \item \textbf{\textit{Manual Correction \& Verification against $\mathcal{V}$:}} 
        Human verifiers receive the video $\mathcal{V}$, question $\mathcal{Q}$, and correct gold answer $\mathcal{A}$ to ensure proper grounding of the causal chain in the video content. Chains passing the first two stages are checked for logical coherence, relevance, and consistency with $\mathcal{Q}$, $\mathcal{A}$, and $\mathcal{V}$. Verifiers—computer science graduates familiar with the task—may add missing details to causal chains and flag hallucinations (\eg, events not observable or implied in the video). Chains with hallucinations or other failures are regenerated until they pass all checks.
    \end{enumerate}
\end{enumerate}

Only the chains that have passed all the checks are considered for the final dataset. Finally, 1000 samples belonging to each base dataset are manually reviewed by the authors. Over 95\% of the chains passed the author verification, ensuring accurate causal chain annotations. 
Our method produced reliable, accurate causal chains grounded in the video context. In total, we constructed human-verified causal chains for 46,024 samples across three datasets. Further stats provided in the \textcolor{RoyalBlue}{Appendix}. \textcolor{RubineRed}{Dataset will be released for reproducibility and future research.}

%% file: Sections/Approach.tex
\section{Approach}
\label{sec:approach}

\subsection{Overview and Motivation}

Conceptually, our work is grounded in the principles of Structural Causal Models (SCMs) \cite{pearl2009causality} and Chain-of-Thought (CoT) \cite{wei2022chain} reasoning. We decompose the VideoQA task into two explicit stages: \textbf{1)} causal chain extraction; and \textbf{2)} causal chain-driven answering. 
Please refer to \autoref{fig:concept}. From an SCM perspective, we introduce causal chains as structured intermediate representations. While many existing VideoQA approaches can be abstracted as an undifferentiated model where video ($\mathcal{V}$) and question ($\mathcal{Q}$) jointly influence the answer ($\mathcal{A}$) as ($\mathcal{V}\rightarrow \mathcal{A} \leftarrow \mathcal{Q}$), we propose a \textit{structured reasoning via a Causal Reasoning Trace} ($\mathcal{C}$), yielding the following SCM:
$\mathcal{V}\rightarrow \mathcal{C} \rightarrow \mathcal{A}, \mathcal{Q}\rightarrow \mathcal{C} \rightarrow \mathcal{A}$.
Then, we \textit{factorize} this SCM into: 
$\mathcal{V}\rightarrow \mathcal{C} \leftarrow \mathcal{Q}$ (this becomes our \textit{Causal Chain Extractor}---\autoref{fig:concept}(3)) and $\mathcal{Q}\rightarrow \mathcal{A} \leftarrow \mathcal{C}$ (this becomes our \textit{Causal Chain-driven Answerer}---\autoref{fig:concept}(4)). Causal Chain Extractor module would first extract $\mathcal{C}$, and pass it to the Causal Chain-Driven Answerer module.

We assume access to fully observed videos, where the causal process leading to an outcome manifests as a temporally ordered sequence of events—a single realized traversal through the underlying causal graph. Once this process is complete, the relevant causal information resides entirely within this factual path. Thus, for Causal-Why VideoQA, the causal chain can be modeled as a linear sequence of observed cause–effect events, providing a complete \& interpretable representation of the reasoning process. Detailed treatment of the proposed use of linear causal chains is provided in the \textcolor{RoyalBlue}{Appendix}.
The structured factorization aligns with CoT principles by making reasoning steps explicit, \& reflects SCM modularity by separating causal understanding from answering. As a result, our approach enables:
\begin{itemize}[wide, labelindent=0pt]
    \item \textbf{Focused processing:} 
    Unlike existing VideoQA methods, our structured approach enables focused processing, with each stage’s output passed to the next. Processing smaller chunks reduces the risk of missing reasoning steps or making incorrect inferences. It is inspired by the Chain-of-Thought philosophy, adapted into a structured model.
    \item \textbf{Improved video understanding:} Modeling cause-and-effect relationships explicitly in causal chains is a dense prediction task. With detailed supervision, vision models learn to better capture rich video content. In contrast, typical VideoQA models are trained with a single label, which may be insufficient to capture the video’s complexity \cite{bi2025reasoning}.
    \item \textbf{Enhanced explainability:} By generating human-readable causal chains as intermediate outputs, our model improves both explainability \& interpretability, making the reasoning process more transparent.
\end{itemize}

\subsection{Model} 
Thus far, we saw that our approach factorizes VideoQA problem into modules: causal chain extraction and causal chain-driven answerer. Now, we explain these individual modules and their training/inference in the following.

\subsubsection{Causal Chain Extractor (CCE)}
The objective of CCE module is to explicitly capture the cause-and-effect relationships or the reasoning steps from videos, conditioned on the questions, and express them in a detailed yet concise form.
The CCE extracts the causal chain after the entire video has been observed, when the full causal process and its outcome are known. At this stage, the causal structure is realized as a single factual sequence of events leading to the answer. The extractor therefore identifies this linear causal path directly, without reasoning over counterfactual or hypothetical alternatives. This formulation treats causal chain extraction as a descriptive recognition task, emphasizing interpretability, fidelity, and robustness within the VideoQA paradigm.
Generating causal chains from videos conditioned on questions is a complex reasoning task. We leverage the representational capacity of foundation models. However, generating causal chains is non-trivial and beyond existing vision-language foundation models. Although foundation models exhibit strong performance across vision-language benchmarks, they lack structured causal supervision, limiting causal chain generation. To address this limitation, we introduce a novel dataset of video–question–causal chain triplets (\autoref{sec:dataset}). This dataset enables supervised fine-tuning (SFT) of pretrained vision-language models for structured causal inference. This trains the model to capture event-level dependencies and generate human-interpretable causal chains grounded in visual and linguistic context.
Formally, given a video $\mathcal{V}$ and a corresponding question $\mathcal{Q}$, the model learns a mapping: $f_{CCE}: (\mathcal{V}, \mathcal{Q}) \rightarrow \mathcal{C}$; where $\mathcal{C}$ denotes the generated causal chain and $f_{CCE}$ represents the fine-tuned vision-language foundation model. Further model implementation details provided in \textcolor{RoyalBlue}{Appendix}.

\begin{figure}
    \centering
    \includegraphics[width=0.9\columnwidth]{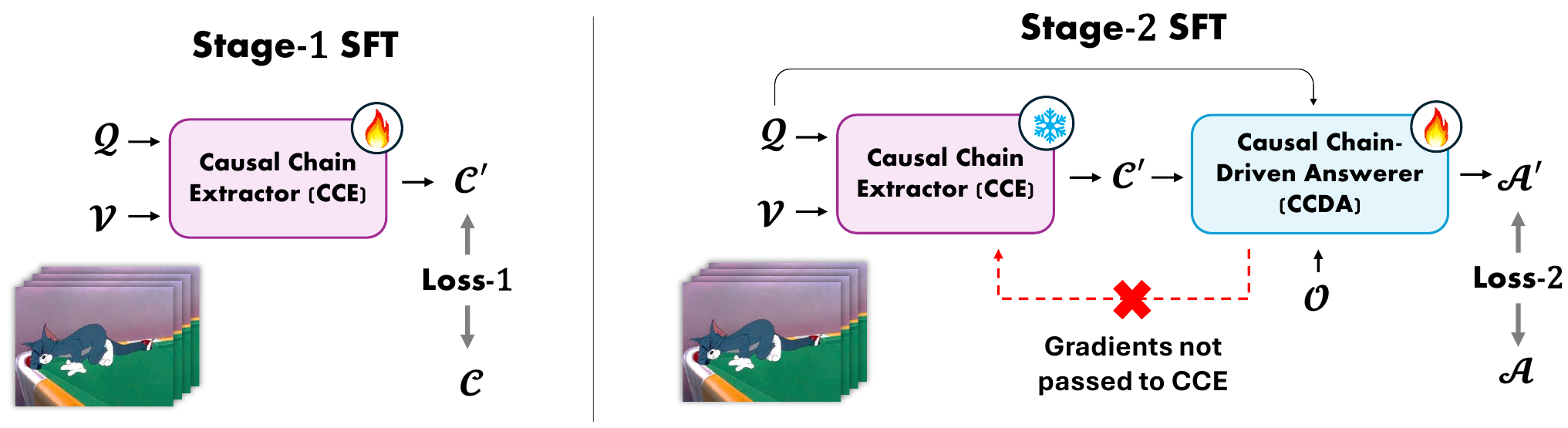}
    \caption{\textbf{Stage-wise training of our model.}}
    \label{fig:model_training}
\end{figure}

\subsubsection{Causal Chain-Driven Answerer (CCDA)} 
The objective of CCDA module is to select the correct answer from candidate answers, based on the questions and the causal chains. Existing large language models have shown strong performance on a variety of language-based tasks. We propose leveraging language models to implement the CCDA. However, processing causal chains using language models is non-trivial and mostly not covered in their training suites, largely due to lack of explicit causal chains or step-by-step annotations of causal reasoning. Towards that end, we resort to supervised finetuning of CCDA. Specifically, CCDA receives the extracted causal chain, question, and candidate answer options (typically four to five in existing datasets) and is prompted to select the correct answer. Notably, the Causal Chain Extractor (CCE) remains frozen at this stage. 
Formally, given a question $\mathcal{Q}$, the corresponding causal chain $\mathcal{C}$, and answer options $\mathcal{O}$ the model learns a function: $f_{CCDA}: (\mathcal{Q},\mathcal{C},\mathcal{O}) \rightarrow \mathcal{A}$; where $\mathcal{A}$ denotes the selected answer option and $f_{CCDA}$ represents a finetuned language model.

\subsubsection{Clean Separation with Stage-wise Training}
\noindent\textbf{Problem with end-to-end (E2E) training.} 
E2E VQA models optimize all components—from feature extraction to answer prediction—using final answer loss. This causes gradient leakage, letting error signals flow through the entire pipeline. As a result, models may learn ``shortcuts" that boost final answer accuracy but harm  causal reasoning.\\
\noindent\textbf{Our solution---stagewise training.} Our two-stage, stagewise training prevents this by introducing a clear boundary:
\begin{itemize}
    \item \textit{CCE Training:} 
    The CCE is trained independently to match the ground-truth causal chains ($C$), using a loss defined on $C$ that enforces accurate causal grounding.
    \item \textit{CCDA Training:} 
    The CCDA is trained on the CCE’s output, with the CCE’s weights typically frozen or its predictions treated as fixed during this stage.
\end{itemize}
By freezing the CCE, no gradients can pass backward from the CCDA's answering loss to the CCE's internal weights. This ensures that the CCE's learned causal reasoning logic remains pure and untainted by the pressure to maximize the answer score, thus preserving the integrity, robustness, and causal semantics of the intermediate representation.

\subsubsection{Inference} 
During inference, the CCE predicts a causal chain from the video and question \textit{without} using ground-truth chains. The CCDA then uses this chain, the question, and answer options to select the correct answer.

\subsection{\textbf{\textsc{CauCo:}} A Causal Coherence Metric}
We introduce the CauCo score to evaluate how well generated causal chains model causality—an aspect regular captioning metrics miss. CauCo quantifies causal consistency and provides causal guarantees in open-world settings by measuring whether events are logically linked by cause and effect.
Following the LLM-as-verifier approach, we SFT an LLM to judge whether a chain is causally coherent. The evaluator outputs “True” or “False.” For SFT, we construct positive and negative samples: positive ones are correct causal chains, and negatives are created by perturbing them using six strategies—\textit{actor swapping}, \textit{event negation}, \textit{event removal}, \textit{event order reversal}, \textit{semantic modification}, and \textit{chain shuffling}.

%% file: Sections/Experiments.tex
\section{Experiments}
\label{sec:experiments}

This section presents experiments validating our hypotheses. We first establish an upper bound for causal VideoQA using our method, then compare it to state-of-the-art approaches, and finally assess explainability through human studies. The experimental setup is outlined below.

\noindent\textbf{Implementation details.} We use PyTorch \cite{paszke2019pytorch} to implement models. Noting the strong performance of VILA 1.5 and LLaMA on vision and language tasks, we adopt them as representative models in implementing our CCE and CCDA, respectively. Note that our approach is not designed for or limited to any models; practitioners may use models of their choice such as \cite{videollama, videochat2, vila, internvl25}. Our CCE is based on VILA-3B \cite{vila}; CCDA is LLaMA-3.1-8B \cite{grattafiori2024llama} version. Further implementation details are provided in \textcolor{RoyalBlue}{Appendix}. \textcolor{RubineRed}{Codebase along with causal chains will be released.}

\noindent\textbf{Datasets.} We conduct experiments on the datasets as discussed in \autoref{sec:dataset}.

\noindent\textbf{Tasks.} 
We evaluate on a multiple-choice QA task with five answer options per question, only one of which is correct.

\noindent\textbf{Performance metric.} Following prior work \cite{nextqa, causalvidqa, causalchaos}, we use Accuracy as the performance metric. For causal chain quality, we use our CauCo score and captioning metrics: BLEU \cite{bleu}, Meteor \cite{meteor}, ROUGE \cite{rouge}, SPICE \cite{spice}.

\subsection{Experimental Upper bound on Performance}

\input{Tables/results_theoretical_ub}

We first investigate whether causal chains can serve as effective intermediate representations. To this end, we use groundtruth causal chains instead of predicted ones, \textit{training \& testing} the Causal Chain-Driven Answerer on these annotations. This setting simulates an ideal scenario where causal chain generation is perfectly accurate. Thus, this experiment aims to determine the model’s performance upper bound when provided with flawless causal chains.

The results, summarized in \autoref{tab:res_theoretical_ub}, show near-perfect accuracy, even surpassing human performance. This finding suggests that using causal chains as intermediate representations is a promising paradigm worth further exploration.

\begin{figure}
    \centering
    \includegraphics[width=\columnwidth]{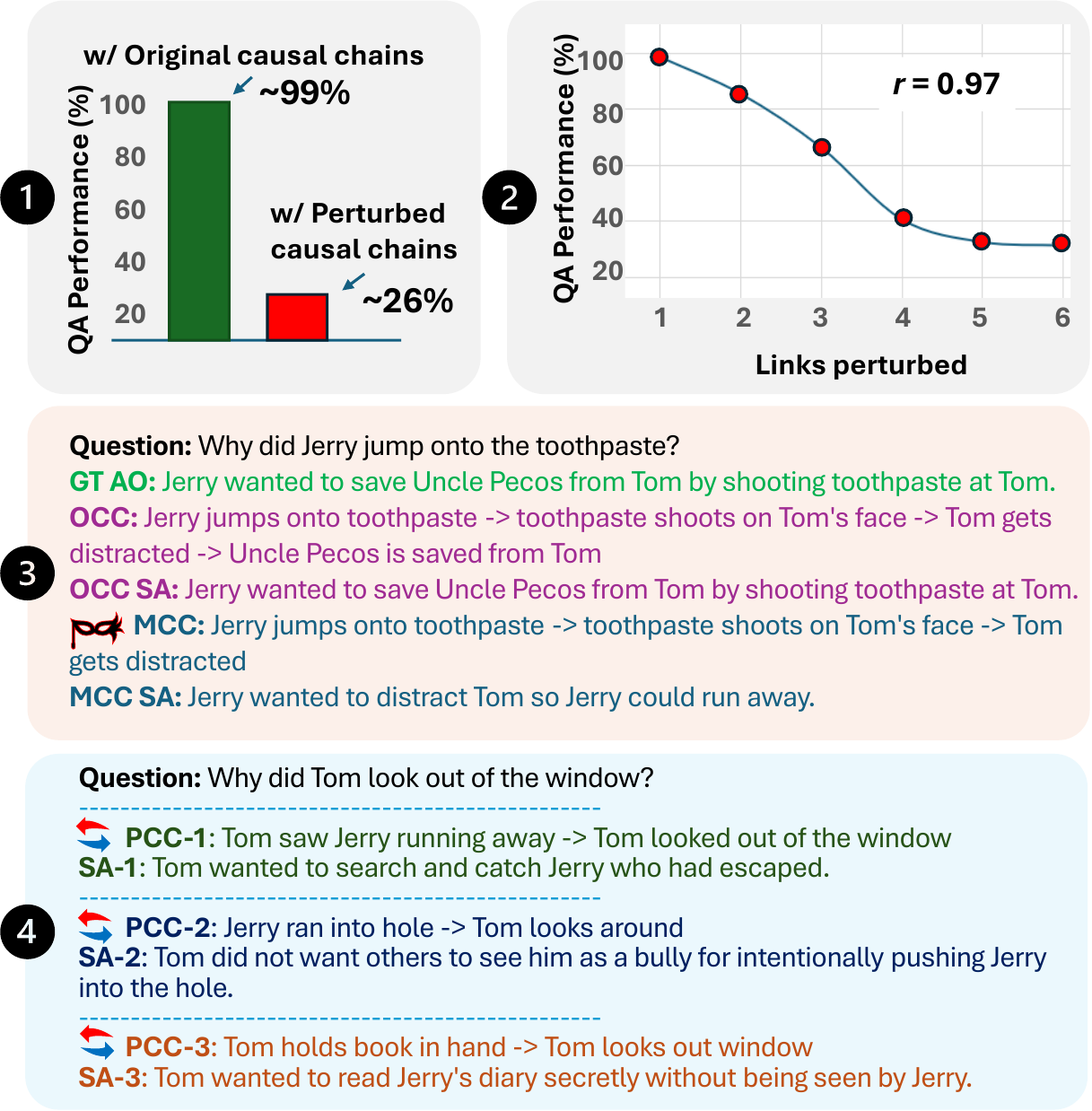}
    \caption{\textbf{Causal chain ablation study.} \textbf{(1)} Study-I: QA accuracy drops by 73\% when chains are perturbed. \textbf{(2)} Study-II: drop in QA is correlated to amount of perturbation. \textbf{(3,4)} Qualitative example. OCC: original causal chains, MCC: masked chains, SA: selected answer, PCC: perturbed chains. SA changes intuitively as chains are perturbed. \textit{Please zoom-in.}}
    \label{fig:causal_chain_sensitivity}
\end{figure}

\subsection{Ablation Studies on the Role of Causal Chains}
To test whether CCDA’s improvements arise from genuine causal reasoning rather than surface-level context enrichment, we conducted ablation studies that systematically degrade the quality of causal chains. 
\textbf{Study-I:} We semantically perturbed the causal chains such that contextual information was preserved but causal relations were disrupted. We observed a sharp decline in QA accuracy once causal information was perturbed, despite the context remaining intact (\autoref{fig:causal_chain_sensitivity}(1)). 
\textbf{Study-II:} 
We progressively masked links and measured QA performance; accuracy degraded monotonically with increased perturbation (\autoref{fig:causal_chain_sensitivity}(2)). Causal-chain quality strongly correlated with CCDA accuracy (\textit{r}=0.97).
These results empirically validate that CCDA’s reasoning depends on causal-chain integrity, with improvements driven by causal reasoning rather than spurious correlations.
Qualitative examples (\autoref{fig:causal_chain_sensitivity}(3,4)) show how chain perturbations intuitively alter CCDA’s answers.

\subsection{Performance Comparison with SOTA}

\paragraph{Baselines.} Following prior work \cite{nextqa, causalvidqa, causalchaos}, we compare against a wide range of models: \textbf{1)} \textit{traditional approaches} \cite{evqa, comem, hme, hcrn, hga};  
\textbf{2)} \textit{causal approaches} \cite{wei2023visual,zang2023discovering};  
and \textbf{3)} \textit{SOTA VLMs/Multimodal foundation approaches} \cite{mist, blip2, videollama, videochat2, gpt4o, vila, deepseekvl2, Qwen2.5-VL}, 
which excel on vision-language tasks. 

\input{Tables/results_perf_overall}

\begin{figure*}
    \centering
    \includegraphics[width=\textwidth]{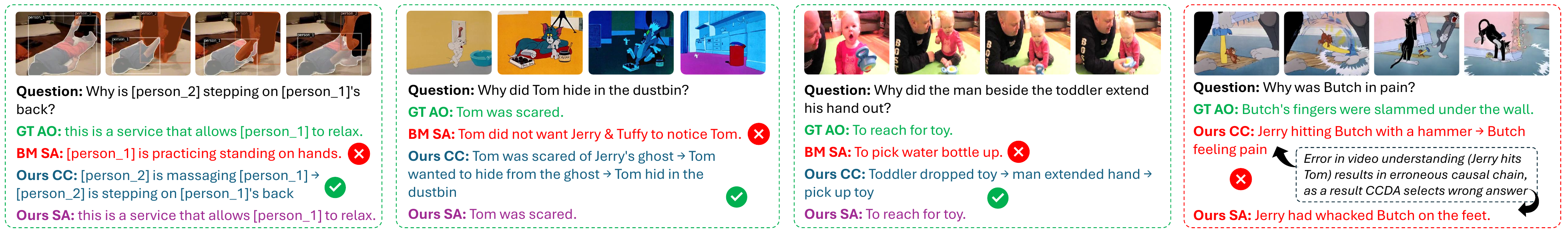}
    \caption{\textbf{Qualitative results.} 
    GT AO: Groundtruth Answer Option; BM: Baseline Model; SA: Selected Answer; CC: Causal Chain. Only a few frames per video are shown. Green and red boxes indicate success and failure cases. In the first example, actor masks come from the CausalVidQA dataset, which includes reference-based QA. \textit{Please zoom-in.}}
    \label{fig:qual_res}
\end{figure*}

\paragraph{Results.} 
\autoref{tab:res_sota} reveals that traditional VideoQA models lag behind VLMs. MIST, a smaller model, performs well—likely from its long-term video understanding and use of CLIP features with spatiotemporal attention. Our method further outperforms prior causal models such as MCR+HCRN \cite{zang2023discovering}, the closest to our work, while remaining simpler and producing interpretable causal chains, unlike latent-variable or counterfactual methods.
Next-generation VLMs such as VILA \cite{vila} outperform GPT-4o on causal VideoQA. Building on VILA, our framework surpasses prior models through improved visual grounding via causal chain extraction and explicit CCDA reasoning. In contrast, VLMs perform grounding and reasoning jointly, often overlooking visual evidence and relying on plausibility biases. Our structured model guides the extractor to capture cause–effect relations in videos—a dense prediction task enhanced by fine-grained supervision. Typical VideoQA models use a single-label objective, leading to shallow learning, whereas our approach isolates comprehension from bias, allowing the CCDA to focus on selecting answers from extracted causal chains. The CCDA made fewer errors, though the extractor occasionally misinterpreted object roles or actions, producing flawed chains and reducing accuracy (\autoref{fig:qual_res}).
Based on our analysis, future work should focus on \textbf{1)} role or relationship modeling and \textbf{2)} situation understanding. We also compare our model’s causal chains with those from QwenVL2.5, a representative SOTA VLM (\autoref{tab:cc_gen_res}, averaged across datasets). Our model performs significantly better, suggesting that causality is often overlooked in multimodal understanding \cite{ma2025causal}—a gap our work addresses.
We believe models like ours also have the potential to serve as reasoning engines, which should be explored by future work. 

\input{Tables/results_causalchain_gen}

\subsection{Human Studies}
Explainable systems benefit two user groups: researchers/ system designers \& consumers. We design four studies to evaluate our explainable system against black-box models across multiple criteria. Since most SOTA VLMs are black boxes, we use VILA \cite{vila} as a representative model for its strong performance \& lack of explicit explanations. 
To ensure reliability, we conducted user studies with six participants, exceeding the sample size used in prior work \cite{causalchaos}.
To minimize bias: 1) no study participants were involved in the project; 2) Sample order randomization \& blinding (independent, anonymized tests) was used for all participants.
Further details provided in \textcolor{RoyalBlue}{Appendix}.

\subsubsection{Study I: Explainability}
In this study, participants are presented with 50 questions and models' outputs, including final answers and, when available, intermediate explanations in the form of causal chains. They rate their understanding of the explanation,\ie, causal chains, regardless of the correctness of the final answer. To avoid the ambiguity of arbitrary scales (\eg, 1–5), we use a comparative evaluation: participants choose between: \textbf{1)} \textit{System A}---no explanation (\textit{BlackBox model}); \textbf{2)} \textit{System B---Our Model} with causal chain explanations; or \textbf{3)} \textit{No Preference}---indicating no added value from the causal chains as explanations. \autoref{tab:res_hs_explainability} shows participants found causal chains useful explanations in over 69\% of cases, demonstrating their overall effectiveness. Nonetheless, in $\sim$29\% of cases, the explanations were not considered helpful, suggesting opportunities for further refinement.

\subsubsection{Study II: Trustworthiness}
\input{Tables/results_hs_consolidated}
In this study, participants are presented with 50 questions and models' outputs, including final answers and, when available, intermediate explanations in the form of causal chains. The goal is to evaluate users' trust in the systems based on the explanations and the perceived correctness of the answers, without access to ground-truth labels. Participants select one of the following options: \textbf{1)} \textit{System A}---trust the \textit{BlackBox} model’s answer; \textbf{2)} \textit{System B}---trust \textit{Our Model’s} answer with its causal chain explanation; or \textbf{3)} \textit{No Preference}---no clear preference between the two. \autoref{tab:res_hs_explainability} summarizes that in over 62\% of cases, participants expressed significantly greater trust in our model. However, trust declined when the model's predictions conflicted with participants' expectations. We anticipate that trust will increase as the accuracy of our approach improves.

\subsubsection{Study III: Human Preference}
In this study, we analyze human preference between the two systems: \textbf{1)} \textit{BlackBox model}; or \textbf{2)} \textit{Our Model}. To ensure a fair comparison, we consider only successful predictions (\ie, those matching groundtruth answers) from both models. Human participants evaluate 50 samples, selecting their preferred model. Participants are instructed to assume \textit{they are using an AI system designed to study human behavior and explain the ``why" behind people's actions in the videos}. The results of this study are presented in \autoref{tab:res_hs_explainability}. We found that in over 85\% of cases, participants strongly preferred a system that provides explanations, like our approach, to support its decisions.

\subsubsection{Study IV: Utility in System Debugging}
\input{Tables/results_hs_rd}
Researchers and system designers seek to improve models through systematic debugging. Identifying limitations is the first step; intermediate outputs such as causal chains alongside final predictions provide useful insights. We tested this by examining failure cases of our explainable and black-box models. While this assumes known errors, it reflects real settings where designers work with labeled data.
Researchers are then asked to identify the source of the model’s failure: \textbf{1)} visual perception, \textbf{2)} language and reasoning, or \textbf{3)} Cannot Tell. The third option indicates uncertainty, while choosing the first two suggests a clearer understanding of the system and supports more confident diagnosis.
Six researchers with expertise in computer vision, VLMs, and LLMs each analyze 20 failure cases from both systems, identifying fault locations. For our model, they view the video, question, causal chain explanations, and final predictions; for the black-box model, only the video, question, and final predictions are provided. Participants are instructed to use all available information in their assessments.
\autoref{tab:res_hs_rd} shows that the "Cannot Tell" option was chosen far less often for our model than in case of BlackBox model. Researchers attributed failures to the chain extractor in $\sim$48\% of cases and to the chain-driven answerer in $\sim$38\%, suggesting that causal chains improve system debuggability.

\subsection{Out-Of-Domain Chain Generation}
\input{Tables/results_ood_cc_gen}

\begin{figure}
    \centering
    \includegraphics[width=0.95\columnwidth]{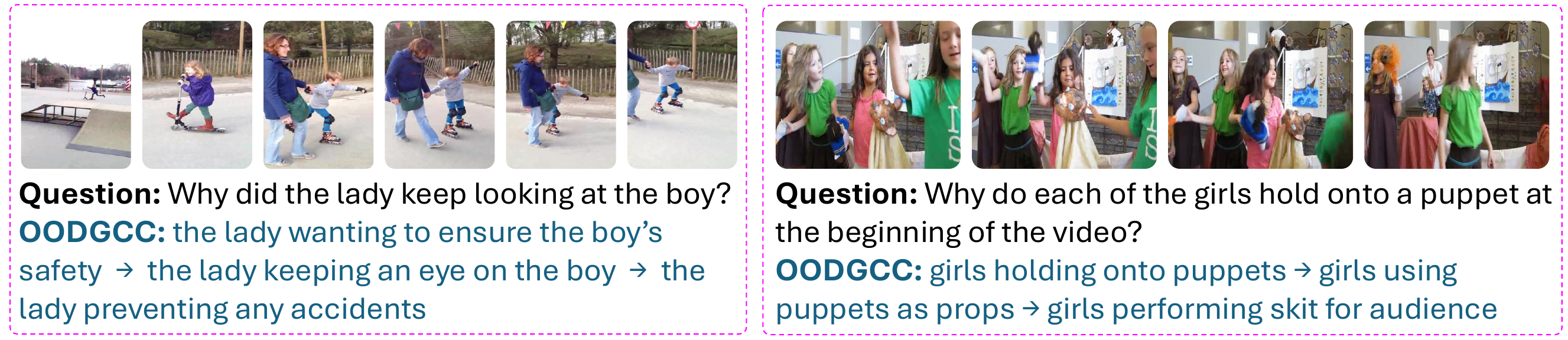}
    \caption{\textbf{Qualitative examples of Out-of-Domain generated causal chains (\textsc{oodgcc}).} \textit{Please zoom-in}.}
    \label{fig:oodgcc_qual_res}
\end{figure}

To test the generalizability of CCE, we train it on a cartoon-based dataset (CausalChaos!) and evaluate it on an out-of-distribution real-world dataset (NextQA), making the task intentionally challenging.
Quality of generated chains is evaluated in terms of \textsc{bleu-1–4 (b), meteor (m), rouge (r), spice (s)}, our causal coherence score \textsc{ccs}, w.r.t. groundtruth causal chains (\autoref{sec:dataset}). A zero-shot \textsc{vila}  1.5 serves as the baseline; our extractor also uses \textsc{vila} 1.5 for a fair comparison. 
\autoref{tab:res_oodcg} shows that our model significantly outperforms the baseline in cross-dataset causal chain generation---suggesting that the reasoning patterns that our causal chain extractor learns are robust and transferable, highlighting its potential as a reusable causal reasoning engine. Qualitative results in \autoref{fig:oodgcc_qual_res}.

%% file: Tables/results_theoretical_ub.tex
\begin{table}[]
\small
\centering
    \resizebox{\columnwidth}{!}{
\setlength\tabcolsep{8pt}
\begin{tabular}{@{}l|cccc@{}}
\toprule
\textbf{Dataset} & \textbf{NextQA} & \textbf{CVQA} & \textbf{CausalChaos!} & \textbf{Overall} \\ \midrule
\textbf{CCDA Accuracy}    &    99.70        &    99.85           &       98.65                &      99.40         \\ \bottomrule
\end{tabular}
} 
\caption{\textbf{Experimental results (Accuracy in \%) for answering based on ground truth causal chains.}}
\label{tab:res_theoretical_ub}
\end{table}

%% file: Tables/results_perf_overall.tex
\begin{table}[]
\centering
\small
\resizebox{1\columnwidth}{!}{
\setlength\tabcolsep{3pt}
\begin{tabular}{@{}lrrccc@{}}
\toprule
\textbf{Model} & \textbf{NeXT-QA}~\cite{nextqa} & \textbf{CausalVidQA}~\cite{causalvidqa} & \textbf{CausalChaos!QA}~\cite{causalchaos} & \textbf{Avg} & \textbf{WtAvg} \\ \midrule
\multicolumn{6}{c}{\textit{Traditional VideoQA approaches}} \\ \midrule
BlindQA \cite{evqa}        & 28.38 & 59.46 & 13.07 & 33.64 & 44.87 \\
EVQA \cite{evqa}          & 42.31 & 60.95 & 13.48 & 38.91 & 50.62 \\
CoMem \cite{comem}          & 46.15 & 62.79 & 13.88 & 40.94 & 53.05 \\
HME \cite{hme}            & 46.52 & 61.45 & 14.02 & 40.66 & 52.43 \\
HCRN \cite{hcrn}           & 47.00 & 61.61 & 17.00 & 41.86 & 52.93 \\
HGA \cite{hga}            & 47.00 & 63.51 & 15.36 & 42.00 & 53.88 \\ \midrule
\multicolumn{6}{c}{\textit{Semi-traditional approaches}} \\ \midrule
MIST \cite{mist}           & 54.79 & 72.41 & 44.88 & 57.36 & 64.04 \\ \midrule
\multicolumn{6}{c}{\textit{Causal modeling approaches}} \\ \midrule
VCSR \cite{wei2023visual}           & 53.00 & 65.41 & -     & -     & -     \\
MCR+HCRN \cite{zang2023discovering}       & 49.20 & 66.00 & -     & -     & -     \\ \midrule
\multicolumn{6}{c}{\textit{VLMs/Multimodal foundation approaches}} \\ \midrule
BLIP-2 \cite{blip2}         & 45.00 & 62.00 & 23.32 & 43.44 & 53.00 \\
VideoLLaMA \cite{videollama}     & 42.00 & 31.00 & 11.73 & 28.24 & 33.31 \\
VideoChat2 \cite{videochat2}     & 60.00 & 46.00 & 15.36 & 40.45 & 48.50 \\
GPT4o \cite{gpt4o}         & 70.00 & 52.00 & 48.17 & 56.72 & 58.00 \\
VILA 1.5-3B \cite{vila}    & 60.23 & 72.11 & 62.80 & 65.05 & 67.20 \\
DeepSeek-VL2 \cite{deepseekvl2}    & 51.55 & 57.08 & 17.12 & 41.92 & 51.95 \\
QwenVL 2.5-7B \cite{Qwen2.5-VL}   & 70.75 & 76.22 & 31.54 & 59.50 & 70.73 \\ \midrule
Ours           & 63.95 & 76.18 & 67.65 & \textbf{69.26} & \textbf{71.22} \\ \bottomrule
\end{tabular}
}
\caption{\textbf{Performance evaluation with SOTA methods.} 
WtAvg: Weighted average.}
\label{tab:res_sota}
\end{table}

%% file: Tables/results_causalchain_gen.tex
\begin{table}[]
\small
\centering
\resizebox{\columnwidth}{!}{
\setlength\tabcolsep{6pt}
\begin{tabular}{@{}lcccccccc@{}}
\toprule
\textbf{Model} & \textbf{B1} & \textbf{B2} & \textbf{B3} & \textbf{B4} & \textbf{M} & \textbf{R} & \textbf{S} & \textbf{CCS} \\ \midrule
QwenVL2.5-3B Oneshot &  0.35 &	0.23 &	0.16 &	0.11 &	0.54 &	0.36 &	0.40 &	0.75
  \\
Ours &  \textbf{0.63} &	\textbf{0.47} &	\textbf{0.36} &	\textbf{0.28} &	\textbf{0.61} &	\textbf{0.50} &	\textbf{0.52} &	\textbf{0.89}
 \\
\bottomrule
\end{tabular}
}
\caption{\textbf{Causal chain generation performance results.}}
\label{tab:cc_gen_res}
\end{table}

%% file: Tables/results_hs_consolidated.tex
\begin{table}
\small
    \centering
    \resizebox{0.95\columnwidth}{!}{
\setlength\tabcolsep{3pt}
    \begin{tabular}{@{}lccc@{}}
    \toprule
        \textbf{Model} & \textbf{Explainability} &  \textbf{Trustworthy} &  \textbf{Human Preferred}\\ \midrule
        Blackbox &01.33 & 01.33 & 14.81\\
        Ours & \textbf{69.33} & \textbf{62.67} & \textbf{85.18}\\
        No Preference & 29.33 & 36.00 & n/a\\ \bottomrule
    \end{tabular}
}    
    \caption{\textbf{Human study results (\%) along various axes.}}
    \label{tab:res_hs_explainability}
\end{table}

%% file: Tables/results_hs_rd.tex
\begin{table}[]
\centering
\small
    \resizebox{0.65\columnwidth}{!}{
\setlength\tabcolsep{8pt}
\begin{tabular}{@{}lccc@{}}
\toprule
\textbf{Model} & \textbf{VLM} & \textbf{LLM} & \textbf{Cannot Tell} \\
\midrule
Blackbox       & 15.00         &15.00          &70.00                 \\
Ours           &  \textbf{48.33}        &\textbf{38.33}        &\textbf{13.33}                 \\ 
\bottomrule
\end{tabular}
}    
\caption{\textbf{Utility from system debugging perspective.}}
\label{tab:res_hs_rd}
\end{table}

%% file: Tables/results_ood_cc_gen.tex
\begin{table}[]
\small
\centering
\setlength\tabcolsep{6pt}
\resizebox{0.95\columnwidth}{!}{
\begin{tabular}{@{}lcccccccc@{}}
\toprule
\textbf{Model} & \textbf{B1}   & \textbf{B2}   & \textbf{B3}   & \textbf{B4}   & \textbf{S}    & \textbf{R}    & \textbf{M} & \textbf{CCS}   \\ \midrule
Baseline & 0.19          & 0.10          & 0.06          & 0.04          & 0.35          & 0.29          & 0.37          & 0.42                  \\
Ours  & \textbf{0.35} & \textbf{0.22} & \textbf{0.13} & \textbf{0.08} & \textbf{0.39} & \textbf{0.37} & \textbf{0.61} &\textbf{0.84} \\ \bottomrule
\end{tabular}
}
\caption{\textbf{\textsc{ood} chain generation performance evaluation.}}
\label{tab:res_oodcg}
\end{table}

%% file: Sections/Conclusion.tex
\section{Conclusion}
\label{sec:conclusion}
We shift paradigms and introduce a principled, structured approach for Causal-Why VideoQA task, which decouples video understanding, reasoning and answer generation. These modules use natural language causal chains as intermediate representations. Our approach enforces analyzing entire videos and explicitly reason to determine the cause behind the actions, instead of shortcut-based answering. At the core of our method lies novel integration of SCMs and CoT. Our approach  We show how to efficiently construct training data to develop such models. Thorough experimentation demonstrates that our approach improves explainabiity, performance on VideoQA task, and have the potential to serve as reusable causal reasoning engines.

%% file: main.bib
@String(CVPR= {IEEE Conf. Comput. Vis. Pattern Recog.})

@String(AAAI = {AAAI})

@String(CVPR  = {CVPR})

@article{videochat2,
  title={Mvbench: A comprehensive multi-modal video understanding benchmark},
  author={Li, Kunchang and Wang, Yali and He, Yinan and Li, Yizhuo and Wang, Yi and Liu, Yi and Wang, Zun and Xu, Jilan and Chen, Guo and Luo, Ping and others},
  booktitle={Proceedings of the IEEE/CVF conference on computer vision and pattern recognition},
  year={2024}
}

@inproceedings{blip2,
      title={{BLIP-2:} Bootstrapping Language-Image Pre-training with Frozen Image Encoders and Large Language Models}, 
      author={Junnan Li and Dongxu Li and Silvio Savarese and Steven Hoi},
      year={2023},
      booktitle={ICML},
}

@inproceedings{videollama,
    title = {Video-{LL}a{MA}: An Instruction-tuned Audio-Visual Language Model for Video Understanding},
    author = {Zhang, Hang  and Li, Xin  and Bing, Lidong},
    editor = "Feng, Yansong  and
      Lefever, Els",
    booktitle = "Proceedings of the 2023 Conference on Empirical Methods in Natural Language Processing: System Demonstrations",
    month = dec,
    year = "2023",
    address = "Singapore",
    publisher = "Association for Computational Linguistics",
    url = "https://aclanthology.org/2023.emnlp-demo.49",
    doi = "10.18653/v1/2023.emnlp-demo.49",
    pages = "543--553",
}

@inproceedings{evqa,
  title={Vqa: Visual question answering},
  author={Antol, Stanislaw and Agrawal, Aishwarya and Lu, Jiasen and Mitchell, Margaret and Batra, Dhruv and Zitnick, C Lawrence and Parikh, Devi},
  booktitle={Proceedings of the IEEE international conference on computer vision},
  pages={2425--2433},
  year={2015}
}

@inproceedings{comem,
  title={Motion-appearance co-memory networks for video question answering},
  author={Gao, Jiyang and Ge, Runzhou and Chen, Kan and Nevatia, Ram},
  booktitle={Proceedings of the IEEE Conference on Computer Vision and Pattern Recognition},
  pages={6576--6585},
  year={2018}
}

@inproceedings{hme,
  title={Heterogeneous memory enhanced multimodal attention model for video question answering},
  author={Fan, Chenyou and Zhang, Xiaofan and Zhang, Shu and Wang, Wensheng and Zhang, Chi and Huang, Heng},
  booktitle={Proceedings of the IEEE/CVF conference on computer vision and pattern recognition},
  pages={1999--2007},
  year={2019}
}

@inproceedings{hcrn,
  title={Hierarchical conditional relation networks for video question answering},
  author={Le, Thao Minh and Le, Vuong and Venkatesh, Svetha and Tran, Truyen},
  booktitle={Proceedings of the IEEE/CVF conference on computer vision and pattern recognition},
  pages={9972--9981},
  year={2020}
}

@inproceedings{hga,
  title={Reasoning with heterogeneous graph alignment for video question answering},
  author={Jiang, Pin and Han, Yahong},
  booktitle={Proceedings of the AAAI Conference on Artificial Intelligence},
  volume={34},
  number={07},
  pages={11109--11116},
  year={2020}
}

@inproceedings{nextqa,
  title={Next-qa: Next phase of question-answering to explaining temporal actions},
  author={Xiao, Junbin and Shang, Xindi and Yao, Angela and Chua, Tat-Seng},
  booktitle={Proceedings of the IEEE/CVF conference on computer vision and pattern recognition},
  pages={9777--9786},
  year={2021}
}

@inproceedings{mist,
  title={MIST: Multi-modal Iterative Spatial-Temporal Transformer for Long-form Video Question Answering},
  author={Gao, Difei and Zhou, Luowei and Ji, Lei and Zhu, Linchao and Yang, Yi and Shou, Mike Zheng},
  booktitle={Proceedings of the IEEE/CVF Conference on Computer Vision and Pattern Recognition},
  pages={14773--14783},
  year={2023}
}

@inproceedings{bleu,
  title={Bleu: a method for automatic evaluation of machine translation},
  author={Papineni, Kishore and Roukos, Salim and Ward, Todd and Zhu, Wei-Jing},
  booktitle={Proceedings of the 40th annual meeting of the Association for Computational Linguistics},
  pages={311--318},
  year={2002}
}

@inproceedings{meteor,
  title={METEOR: An automatic metric for MT evaluation with improved correlation with human judgments},
  author={Banerjee, Satanjeev and Lavie, Alon},
  booktitle={Proceedings of the acl workshop on intrinsic and extrinsic evaluation measures for machine translation and/or summarization},
  pages={65--72},
  year={2005}
}

@inproceedings{rouge,
  title={Rouge: A package for automatic evaluation of summaries},
  author={Lin, Chin-Yew},
  booktitle={Text summarization branches out},
  pages={74--81},
  year={2004}
}

@inproceedings{spice,
  title={Spice: Semantic propositional image caption evaluation},
  author={Anderson, Peter and Fernando, Basura and Johnson, Mark and Gould, Stephen},
  booktitle={Computer Vision--ECCV 2016: 14th European Conference, Amsterdam, The Netherlands, October 11-14, 2016, Proceedings, Part V 14},
  pages={382--398},
  year={2016},
  organization={Springer}
}

@InProceedings{causalvidqa,
    author    = {Li, Jiangtong and Niu, Li and Zhang, Liqing},
    title     = {From Representation to Reasoning: Towards both Evidence and Commonsense Reasoning for Video Question-Answering},
    booktitle = {Proceedings of the IEEE/CVF Conference on Computer Vision and Pattern Recognition (CVPR)},
    month     = {June},
    year      = {2022}
}

@misc{gpt4o,
    author = "{OpenAI}",
    title = "Hello GPT-4o",
    year = "2024",
    howpublished = "\url{https://openai.com/index/hello-gpt-4o/}",
    note = "[Online; accessed 31-May-2024]"
}

@inproceedings{vila,
  title={Vila: On pre-training for visual language models},
  author={Lin, Ji and Yin, Hongxu and Ping, Wei and Molchanov, Pavlo and Shoeybi, Mohammad and Han, Song},
  booktitle={Proceedings of the IEEE/CVF Conference on Computer Vision and Pattern Recognition},
  pages={26689--26699},
  year={2024}
}

@inproceedings{causalchaos,
title={CausalChaos! Dataset for Comprehensive Causal Action Question Answering Over Longer Causal Chains Grounded in Dynamic Visual Scenes},
author={Paritosh Parmar and Eric Peh and Ruirui Chen and Ting En Lam and Yuhan Chen and Elston Tan and Basura Fernando},
booktitle={The Thirty-eight Conference on Neural Information Processing Systems Datasets and Benchmarks Track},
year={2024},
url={https://openreview.net/forum?id=gP4aAi7q8S}
}

@article{wei2022chain,
  title={Chain-of-thought prompting elicits reasoning in large language models},
  author={Wei, Jason and Wang, Xuezhi and Schuurmans, Dale and Bosma, Maarten and Xia, Fei and Chi, Ed and Le, Quoc V and Zhou, Denny and others},
  journal={Advances in neural information processing systems},
  volume={35},
  pages={24824--24837},
  year={2022}
}

@book{pearl2009causality,
  title={Causality},
  author={Pearl, Judea},
  year={2009},
  publisher={Cambridge university press}
}

@article{scholkopf2021toward,
  title={Toward causal representation learning},
  author={Sch{\"o}lkopf, Bernhard and Locatello, Francesco and Bauer, Stefan and Ke, Nan Rosemary and Kalchbrenner, Nal and Goyal, Anirudh and Bengio, Yoshua},
  journal={Proceedings of the IEEE},
  volume={109},
  number={5},
  pages={612--634},
  year={2021},
  publisher={IEEE}
}

@article{narendra2018explaining,
  title={Explaining deep learning models using causal inference},
  author={Narendra, Tanmayee and Sankaran, Anush and Vijaykeerthy, Deepak and Mani, Senthil},
  journal={arXiv preprint arXiv:1811.04376},
  year={2018}
}

@inproceedings{tang2020unbiased,
  title={Unbiased scene graph generation from biased training},
  author={Tang, Kaihua and Niu, Yulei and Huang, Jianqiang and Shi, Jiaxin and Zhang, Hanwang},
  booktitle={Proceedings of the IEEE/CVF conference on computer vision and pattern recognition},
  pages={3716--3725},
  year={2020}
}

@article{li2020causal,
  title={Causal discovery in physical systems from videos},
  author={Li, Yunzhu and Torralba, Antonio and Anandkumar, Anima and Fox, Dieter and Garg, Animesh},
  journal={Advances in Neural Information Processing Systems},
  volume={33},
  pages={9180--9192},
  year={2020}
}

@inproceedings{yang2021deconfounded,
  title={Deconfounded video moment retrieval with causal intervention},
  author={Yang, Xun and Feng, Fuli and Ji, Wei and Wang, Meng and Chua, Tat-Seng},
  booktitle={Proceedings of the 44th international ACM SIGIR conference on research and development in information retrieval},
  pages={1--10},
  year={2021}
}

@article{chen2025mecd,
  title={MECD: Unlocking multi-event causal discovery in video reasoning},
  author={Chen, Tieyuan and Liu, Huabin and He, Tianyao and Chen, Yihang and Ma, Xiao and Zhong, Cheng and Zhang, Yang and Wang, Yingxue and Lin, Hui and Lin, Weiyao and others},
  journal={Advances in Neural Information Processing Systems},
  volume={37},
  pages={92554--92580},
  year={2025}
}

@inproceedings{zang2023discovering,
  title={Discovering the real association: Multimodal causal reasoning in video question answering},
  author={Zang, Chuanqi and Wang, Hanqing and Pei, Mingtao and Liang, Wei},
  booktitle={Proceedings of the IEEE/CVF Conference on Computer Vision and Pattern Recognition},
  pages={19027--19036},
  year={2023}
}

@inproceedings{wang2024modeling,
  title={Modeling Event-level Causal Representation for Video Classification},
  author={Wang, Yuqing and Meng, Lei and Ma, Haokai and Wang, Yuqing and Huang, Haibei and Meng, Xiangxu},
  booktitle={Proceedings of the 32nd ACM International Conference on Multimedia},
  pages={3936--3944},
  year={2024}
}

@inproceedings{ayazoglu2013finding,
  title={Finding causal interactions in video sequences},
  author={Ayazoglu, Mustafa and Yilmaz, Burak and Sznaier, Mario and Camps, Octavia},
  booktitle={Proceedings of the IEEE International Conference on Computer Vision},
  pages={3575--3582},
  year={2013}
}

@article{hanson1955causal,
  title={Causal chains},
  author={Hanson, Norwood Russell},
  journal={Mind},
  volume={64},
  number={255},
  pages={289--311},
  year={1955},
  publisher={JSTOR}
}

@article{trabasso1985causal,
  title={Causal thinking and the representation of narrative events},
  author={Trabasso, Tom and Van Den Broek, Paul},
  journal={Journal of memory and language},
  volume={24},
  number={5},
  pages={612--630},
  year={1985},
  publisher={Elsevier}
}

@article{internvl25,
  title={Expanding performance boundaries of open-source multimodal models with model, data, and test-time scaling},
  author={Chen, Zhe and Wang, Weiyun and Cao, Yue and Liu, Yangzhou and Gao, Zhangwei and Cui, Erfei and Zhu, Jinguo and Ye, Shenglong and Tian, Hao and Liu, Zhaoyang and others},
  journal={arXiv preprint arXiv:2412.05271},
  year={2024}
}

@article{gopnik2004theory,
  title={A theory of causal learning in children: causal maps and Bayes nets.},
  author={Gopnik, Alison and Glymour, Clark and Sobel, David M and Schulz, Laura E and Kushnir, Tamar and Danks, David},
  journal={Psychological review},
  volume={111},
  number={1},
  pages={3},
  year={2004},
  publisher={American Psychological Association}
}

@book{pearl2018book,
  title={The book of why: the new science of cause and effect},
  author={Pearl, Judea and Mackenzie, Dana},
  year={2018},
  publisher={Basic books}
}

@article{grattafiori2024llama,
  title={The llama 3 herd of models},
  author={Grattafiori, Aaron and Dubey, Abhimanyu and Jauhri, Abhinav and Pandey, Abhinav and Kadian, Abhishek and Al-Dahle, Ahmad and Letman, Aiesha and Mathur, Akhil and Schelten, Alan and Vaughan, Alex and others},
  journal={arXiv preprint arXiv:2407.21783},
  year={2024}
}

@inproceedings{su2023language,
  title={Language models are causal knowledge extractors for zero-shot video question answering},
  author={Su, Hung-Ting and Niu, Yulei and Lin, Xudong and Hsu, Winston H and Chang, Shih-Fu},
  booktitle={Proceedings of the IEEE/CVF Conference on Computer Vision and Pattern Recognition},
  pages={4951--4960},
  year={2023}
}

@inproceedings{wei2023visual,
  title={Visual causal scene refinement for video question answering},
  author={Wei, Yushen and Liu, Yang and Yan, Hong and Li, Guanbin and Lin, Liang},
  booktitle={Proceedings of the 31st ACM International Conference on Multimedia},
  pages={377--386},
  year={2023}
}

@article{foss2025causalvqa,
  title={CausalVQA: A Physically Grounded Causal Reasoning Benchmark for Video Models},
  author={Foss, Aaron and Evans, Chloe and Mitts, Sasha and Sinha, Koustuv and Rizvi, Ammar and Kao, Justine T},
  journal={arXiv preprint arXiv:2506.09943},
  year={2025}
}

@inproceedings{chen2023llcp,
  title={LLCP: Learning Latent Causal Processes for Reasoning-based Video Question Answer},
  author={Chen, Guangyi and Li, Yuke and Liu, Xiao and Li, Zijian and Al Suradi, Eman and Wei, Donglai and Zhang, Kun},
  booktitle={The Twelfth International Conference on Learning Representations},
  year={2023}
}

@misc{deepseekvl2,
      title={DeepSeek-VL2: Mixture-of-Experts Vision-Language Models for Advanced Multimodal Understanding},
      author={Zhiyu Wu and Xiaokang Chen and Zizheng Pan and Xingchao Liu and Wen Liu and Damai Dai and Huazuo Gao and Yiyang Ma and Chengyue Wu and Bingxuan Wang and Zhenda Xie and Yu Wu and Kai Hu and Jiawei Wang and Yaofeng Sun and Yukun Li and Yishi Piao and Kang Guan and Aixin Liu and Xin Xie and Yuxiang You and Kai Dong and Xingkai Yu and Haowei Zhang and Liang Zhao and Yisong Wang and Chong Ruan},
      year={2024},
      eprint={2412.10302},
      archivePrefix={arXiv},
      primaryClass={cs.CV},
      url={https://arxiv.org/abs/2412.10302},
}

@article{Qwen2.5-VL,
  title={Qwen2.5-VL Technical Report},
  author={Bai, Shuai and Chen, Keqin and Liu, Xuejing and Wang, Jialin and Ge, Wenbin and Song, Sibo and Dang, Kai and Wang, Peng and Wang, Shijie and Tang, Jun and Zhong, Humen and Zhu, Yuanzhi and Yang, Mingkun and Li, Zhaohai and Wan, Jianqiang and Wang, Pengfei and Ding, Wei and Fu, Zheren and Xu, Yiheng and Ye, Jiabo and Zhang, Xi and Xie, Tianbao and Cheng, Zesen and Zhang, Hang and Yang, Zhibo and Xu, Haiyang and Lin, Junyang},
  journal={arXiv preprint arXiv:2502.13923},
  year={2025}
}

@inproceedings{rawal2024dissecting,
  title={Dissecting Multimodality in VideoQA Transformer Models by Impairing Modality Fusion},
  author={Rawal, Ishaan Singh and Matyasko, Alexander and Jaiswal, Shantanu and Fernando, Basura and Tan, Cheston},
  booktitle={International Conference on Machine Learning},
  pages={42213--42244},
  year={2024},
  organization={PMLR}
}

@article{paszke2019pytorch,
  title={Pytorch: An imperative style, high-performance deep learning library},
  author={Paszke, Adam and Gross, Sam and Massa, Francisco and Lerer, Adam and Bradbury, James and Chanan, Gregory and Killeen, Trevor and Lin, Zeming and Gimelshein, Natalia and Antiga, Luca and others},
  journal={Advances in neural information processing systems},
  volume={32},
  year={2019}
}

@inproceedings{ma2025causal,
  title={Causal Inference with Large Language Model: A Survey},
  author={Ma, Jing},
  booktitle={Findings of the Association for Computational Linguistics: NAACL 2025},
  pages={5886--5898},
  year={2025}
}

@article{niu2025r,
  title={R\^{} 3-VQA:" Read the Room" by Video Social Reasoning},
  author={Niu, Lixing and Li, Jiapeng and Yu, Xingping and Wang, Shu and Feng, Ruining and Wu, Bo and Wei, Ping and Wang, Yisen and Fan, Lifeng},
  journal={arXiv preprint arXiv:2505.04147},
  year={2025}
}

@article{bi2025reasoning,
  title={Why reasoning matters? a survey of advancements in multimodal reasoning (v1)},
  author={Bi, Jing and Liang, Susan and Zhou, Xiaofei and Liu, Pinxin and Guo, Junjia and Tang, Yunlong and Song, Luchuan and Huang, Chao and Sun, Guangyu and He, Jinxi and others},
  journal={arXiv preprint arXiv:2504.03151},
  year={2025}
}

@article{gemini,
  title={Gemini 2.5: Pushing the frontier with advanced reasoning, multimodality, long context, and next generation agentic capabilities},
  author={Comanici, Gheorghe and Bieber, Eric and Schaekermann, Mike and Pasupat, Ice and Sachdeva, Noveen and Dhillon, Inderjit and Blistein, Marcel and Ram, Ori and Zhang, Dan and Rosen, Evan and others},
  journal={arXiv preprint arXiv:2507.06261},
  year={2025}
}
